%% file: emnlp2023.tex
\pdfoutput=1

\documentclass[11pt]{article}

\usepackage{EMNLP2023}

\usepackage{times}
\usepackage{latexsym}

\usepackage[T1]{fontenc}

\usepackage[utf8]{inputenc}

\usepackage{microtype}

\usepackage{inconsolata}

\usepackage{lipsum}
\usepackage{listings}
\usepackage{extarrows}
\usepackage{subfigure}
\usepackage{xspace}
\usepackage{graphicx}
\usepackage{makecell}
\usepackage{ifthen}
\usepackage{xifthen}
\usepackage{wrapfig}
\usepackage{amsthm}
\usepackage{stmaryrd}
\usepackage{threeparttable}
\usepackage{enumerate}
\usepackage{multirow}
\usepackage{booktabs}
\usepackage{titlesec}
\usepackage{mathtools, nccmath}
\usepackage{comment}
\usepackage{amsmath,amsfonts,amssymb}
\usepackage[framemethod=TikZ]{mdframed}
\usepackage{lipsum}
\usepackage{float}
\usepackage{threeparttable}
\usepackage{caption}
\usepackage{tcolorbox}
\usepackage{listings}
\usepackage{upquote}
\usepackage[T1]{fontenc}
\usepackage{xcolor}
\usepackage[symbol]{footmisc}
\usepackage{enumerate}
\usepackage{enumitem}
\usepackage{svg}
\usepackage{tabularx}
\usepackage{wrapfig}
\usepackage{lipsum}
\usepackage{stfloats}

\newcommand{\hide}[1]{}

\title{Revisiting Parallel Context Windows: A Frustratingly\\Simple Alternative and Chain-of-Thought Deterioration}

\author{
  Kejuan Yang\textsuperscript{*}, 
  Xiao Liu\textsuperscript{*}, 
  Kaiwen Men, 
  Aohan Zeng, 
  Yuxiao Dong, 
  Jie Tang \\
  Tsinghua University \\
  \texttt{\{ykj22,liuxiao21,mkw22,zah22\}@mails.tsinghua.edu.cn,}\\
  \texttt{\{yuxiaod,jietang\}@tsinghua.edu.cn}
}

\begin{document}
\maketitle
\begin{abstract}
\input{0_abstract.tex}
\end{abstract}

\renewcommand{\thefootnote}{\fnsymbol{footnote}}
\noindent \footnotetext[1]{Kejuan and Xiao contributed equally.}

\input{1_intro.tex}
\input{2_preliminary.tex}

\input{4_experiment.tex}

\input{6_conclusion.tex}

\section*{Limitations}
The limitations of our experimental considerations are as follows:

Firstly, we currently only evaluate decoder-architecture models for their parallel implementation, with none exceeding 20 billion parameters due to our computational constraints. A more comprehensive analysis should extend to larger models, such as LLaMA 65B, known for powerful understanding and CoT reasoning capabilities, and potentially some bidirectional language models~\cite{du2022glm,raffel2020exploring}. 

Secondly, since LLaMA models employ rotary positional embedding, differing from the absolute positional embedding used by GPT2 in ~\cite{ratner2023parallel}, the enhancement brought by PCW may vary.

Thirdly, our experimental scope was restricted to knowledge-intensive tasks like HotpotQA and did not extend to mathematical tasks such as GSM8K~\cite{cobbe2021gsm8k}, which necessitates multi-step reasoning to solve grade-school math word problems. 
We will include more CoT tasks in the next version evaluation.

Therefore, the degradation phenomenon on reasoning tasks caused by parallel windows still requires further exploration and validation. 

\bibliography{anthology,custom}
\bibliographystyle{acl_natbib}

\clearpage
\appendix
\input{appendix.tex}

\end{document}

%% file: 0_abstract.tex
We identify two crucial limitations in the evaluation of recent parallel-integrated method Parallel Context Windows (PCW)~\cite{ratner2023parallel}, which extends the maximum context lengths of language models, e.g., 2048 for LLaMA, by harnessing window-wise attention and positional embedding techniques.
We first show that a simple yet strong baseline, weighted sum ensemble, is missing for the in-context few-shot classification. 
Moreover, on more challenging Chain-of-Thought (CoT) reasoning (e.g., HotpotQA), PCW would present unexpected deterioration regarding question miscomprehension and false inference.
Based on our findings, we suggest that the existing PCW design may not guarantee sufficient improvement and practicality in handling lengthy documents in real-world applications.
More community efforts on enabling language models' long context understanding ability should be paid.

\hide{
Digesting, summarizing, and understanding lengthy documents necessitates a long context window. This capacity, however, is considerably constrained by the maximum token length of Language Models, e.g., 2048 for LLaMA.
Recently, parallel-integrated methods such as Parallel Context Window (PCW)~\cite{ratner2023parallel} suggest adaptions to the attention matrix and positional embedding of demonstration instances to allow for extended context inputs without additional tuning. 
However, our further exploration indicates that PCW simply functions as a parallel majority vote ensemble in classification tasks. Worsely, it deteriorates the CoT reasoning of LMs, leading to question miscomprehension and false inference in the challenging multi-hop reasoning task HotpotQA.
Based on our findings, the parallel-integrated methods do not guarantee coherent and reasonable generation in an extended horizon, thereby casting doubts on their potential applicability in handling intricate and human-like tasks. 
}

%% file: 1_intro.tex
\section{Introduction}

Over the past few months, the field of Large Language Models (LLMs)~\cite{GPT3,chowdhery2022palm,zhang2022opt,scao2022bloom,zeng2022glm} has undergone a remarkable resurgence, primarily GPT-4, which has proved reasoning abilities akin to human, spanning a variety of professional fields from law to mathematics and physics~\cite{openai2023gpt4}.
LLMs experience a paradigm shift, from individual tasks such as machine translation~\cite{lopez2008statistical}, text summarization~\cite{allahyari2017text}, and information extraction~\cite{sarawagi2008information}, and gravitate toward a unified solution where users engage and interact in dialogues with chatbots to query anything. 

\input{figures/fig1.tex}

\input{tables/error_type_table}

Still, a major challenge remains in LLMs --- their abilities are constrained by their maximum context lengths.
For example, GPT-3~\cite{GPT3} mentions its few demonstration samples in in-context learning (ICL) due to length limit.
Recent Auto-GPT~\cite{significant2023autogpt} is also observed to suffer from lengthy histories induced by CoT~\cite{wei2022chain}, which shepherds the LMs to mirror human cognition through a step-by-step progression of thinking and reflection to solve challenging reasoning missions.
Hence it is vital to develop techniques to extend the context length of existing LLMs for reasoning.

A recent related attempt is PCW~\cite{ratner2023parallel}, which brings the idea of parallel contexts to mitigate the length limitation problem in GPTs. 
PCW segments the text sequence into windows, constraining the attention to be visible within each window while all windows share the same positional embeddings.
It reports improvements in few-shot ICL classification and generation tasks over the conventional sequential baseline, especially on fine-grained classification tasks with large label space such as BANKING77~\cite{Casanueva2020} and CLINIC150~\cite{larson-etal-2019-evaluation}.
By introducing over-length number of demonstration samples in one sequence, LMs can access more labels from context and thus outperform the sequential ICL where fewer samples could be seen.

However, in this work we identify limitations in PCW's evaluation, especially from two aspects:
\begin{itemize}[leftmargin=1.5em,itemsep=0pt,parsep=0.2em,topsep=0.0em,partopsep=0.0em]
    \item \textbf{Unequal Comparison}: As PCW sees more demonstrations, it is better to compare sequential methods receiving equal number of samples (e.g., ensembling multiple sequences) instead of a single sequence with fewer samples.
    \item \textbf{Unchallenging Tasks}: PCW evaluates on traditional classification and generation tasks only, but leaves untouched more challenging and practical problems in current LLMs concerning lengthy context of CoT reasoning.
\end{itemize}

\vspace{-1mm}
\paragraph{Contributions.}
In light of the current limitations, we re-examine PCW's effectiveness in few-shot text classification against a fairer baseline and in more challenging CoT problems.

For text classification, we introduce a simple yet strong alternative---Parallel Ensemble (PE), which directly ensembles predictions from each context window as individual sequences, to achieve the same improvement as PCW, without modifying transformers and adding computation complexity (Cf. Figure~\ref{fig1}). 
Results show that PE achieves comparable and even better average performance to PCW in its evaluation.
For more challenging missions, we follow ReAct~\cite{yao2023react} setting to evaluate pure CoT reasoning on closed-book HotpotQA.
Unfortunately, PCW makes no improvement, and even deteriorates LMs CoT reasoning (Cf. Figure~\ref{fig1}).
Careful investigation unveils that PCW might weaken LMs' language reasoning, yielding issues including false inference, question misunderstanding, and absence of CoT (Cf. Figure~\ref{fig2}).

In conclusion, our contributions are two-fold. 
Firstly, we propose that Parallel Ensemble, a direct weighted-sum ensemble on the logits of generated labels, is comparable to PCW on most classification benchmarks without any architecture modification.
Secondly, we examine that PCW unintentionally results in a decline in LM's reasoning ability, raising questions about its practical benefit to current chat-based LLMs.
We appeal to the community for more comprehensive study on the problem of LLMs' length extension challenge.

%% file: figures/fig1.tex
\begin{figure}[t]
  \centering
  \includegraphics[width=0.8\linewidth]{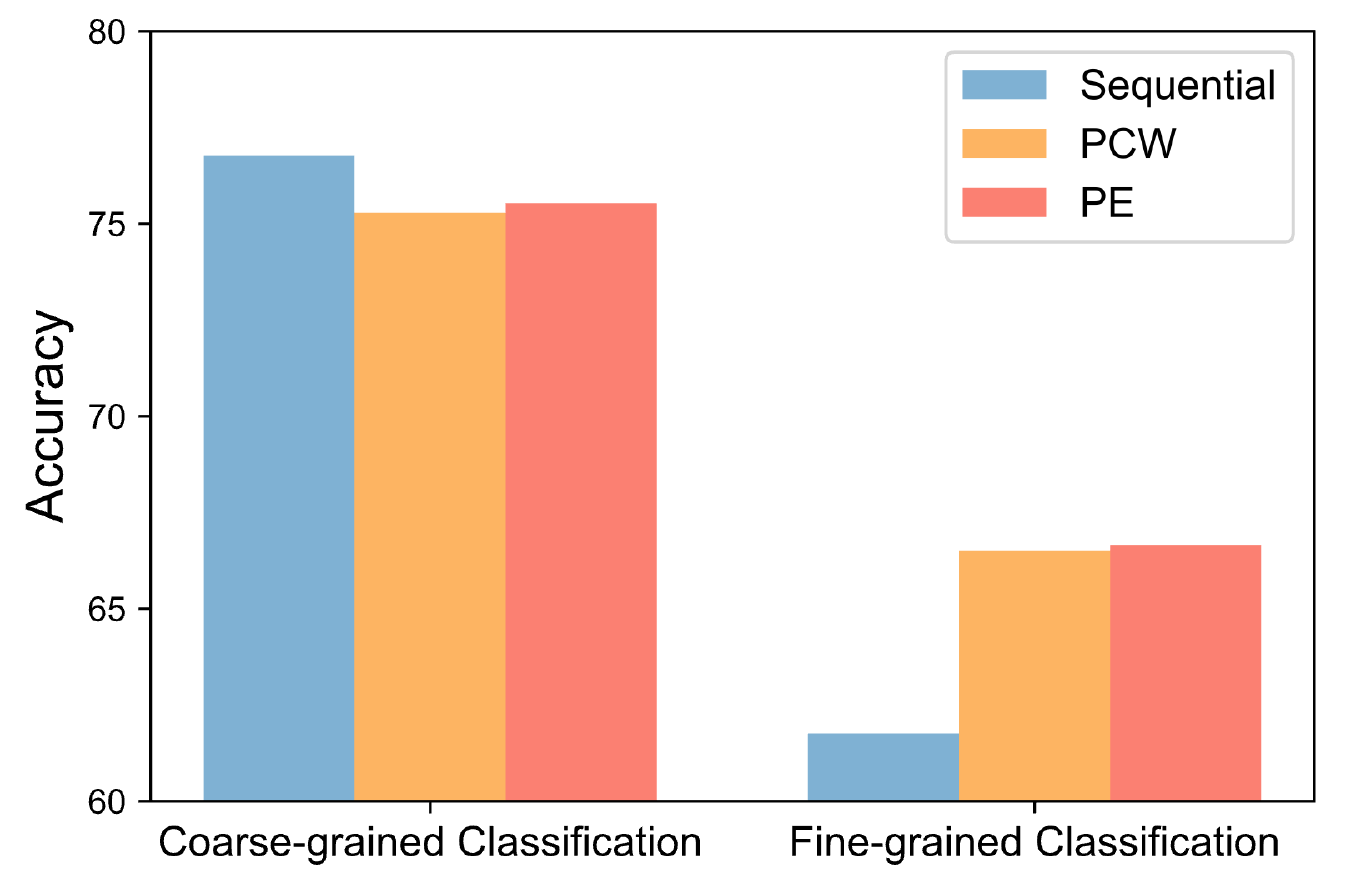}
  \includegraphics[width=0.8\linewidth]{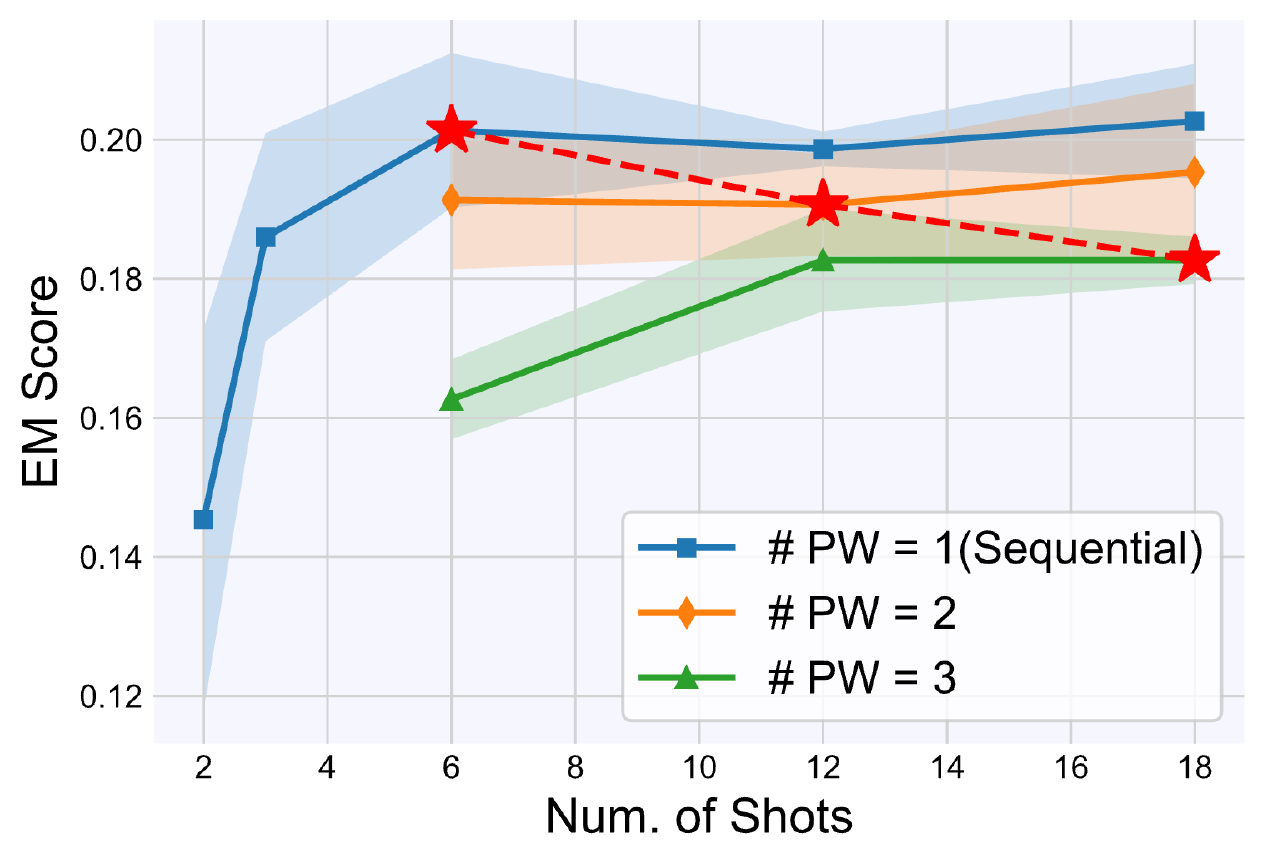}
  \caption{(a) PCW is comparable with Parallel Ensemble~(PE) and outperforms on fine-grained classification benchmarks with over 15 labels;
  (b) PCW deteriorates closed-book HotpotQA.
  The red dashed line illustrates degradation in this challenging multi-hop reasoning task, despite doubling or tripling the number of demonstrations. An increased number of parallel windows (higher \#PW) leads to sparser attention but worse accuracy, while a single window indicates the sequential baseline.}
  \label{fig1}
  \vspace{-5mm}
\end{figure}

%% file: tables/error_type_table.tex
\begin{table}[]
\centering
\resizebox{0.47\textwidth}{!}{%
\begin{tabular}{@{}lcc@{}}
\toprule
\textbf{Error Type}                 & \textbf{Sequential} & \textbf{Parallel} \\ \midrule
Reasoning Error            & 16.28\%                   & 34.09\%              \\
\quad - False Reasoning            & 2.33\%              & 10.23\%           \\
\quad - Question Misinterpretation & 10.47\%             & 19.32\%           \\
\quad - No CoT Reasoning           & 3.49\%              & 4.55\%            \\ \midrule
Non-reasoning Error       & 81.40\%             & 59.09\%              \\ \midrule
Other                      & 2.33\%              & 6.82\%            \\ \bottomrule
\end{tabular}
}
\caption{Analysis on closed-book HotpotQA errors. We classify them into five sub-categories and record their frequencies. 
PCW diminishes reasoning by more false reasoning, misinterpretation of the question, and even a complete lack of CoT reasoning. }
\label{tab:error}
\vspace{-4mm}
\end{table}

%% file: 2_preliminary.tex
\section{Preliminary}
\input{tables/class_table}

\subsection{In-Context Learning}
A language model $\phi$ is pre-trained to predict the conditional probability $p_{\phi}(\psi|C)$ where $C$ represents the text input and $\psi$ represents the word distribution over the given vocabulary. 

In addition to the direct zero-shot inference, LMs also exhibit in-context learning capabilities where they tailor to corresponding tasks by seeing demonstrations(examples). 
In few-shot inference, $C$ is extended into two parts: N-shot demonstrations $D = \{d_1, d_2, ..., d_N\}$ formatted as $d_i = \{input:x_i; output:y_i\}$, and the test input $x_{test}$. 
Conceptually, in-context learning equates to the text generation of $p_{\phi}(y_{test}|D, x_{test})$.

\subsection{Sequential ICL}
The language model reads context input $I = \{T, A, P\}$, which includes text tokens $T$, attention matrix $A$, and positional embedding $P$. 
\begin{itemize}[leftmargin=*,itemsep=0pt,parsep=0.2em,topsep=0.0em,partopsep=0.0em]
    \item Text tokens $T$: tokenized input text.
    \item Attention matrix $A$: a two-dimensional matrix that determines the visibility between input and output tokens---$A_{i,j} = 1$ suggests the $j$-th output token relates to the $i$-th input token, and $A_{i,j} = 0$ suggests no attention between them. 
    \item Positional Embedding $P$: a sequence of IDs indicating the position for every text token. 
\end{itemize}

Denote input token length $l = len(C)$. The standard sequential ICL input $I_{seq}$ is formed as:

\begin{equation}
\label{equation 1}
\fontsize{10pt}{12pt}\selectfont
\begin{aligned}
T_{\text {seq }} & =\left\{T\left(x_{\text {test }}\right), T\left(d_1\right), \cdots, T\left(d_N\right)\right\}, \\
A_{\text {seq }} & = [a_{ij}]_{l \times l} =
    \begin{cases}
      0 & \text{for } 0 \leq j < i < l \\
      1 & \text{otherwise}
    \end{cases}, \\
P_{\text {seq }} & =\{0,1, \cdots, l-1\}.
\end{aligned}
\end{equation}

\subsection{Parallel ICL}
Parallel ICL reconfigures two fundamental inputs of LMs: the attention matrix $A$ and positional embedding $P$.
All demonstrations $D$ are segmented into separate windows $\{W_1, W_2, ...,W_\phi\}$ ~\cite{ratner2023parallel}, denoting the number of windows as $\phi$, where $\phi=N$ is the most fine-grained division. The straightforward parallel approach is to block attention between demonstration windows, but allow the test input $x_{test}$ to attend to every window. 
For positional embedding, we modify the test input to begin after the longest window's position $p_{\text {max}}$. 

The input of Parallel ICL $I_{prl}$ is formulated as: 
\begin{equation}
\label{equation 2}
\fontsize{10pt}{12pt}\selectfont
\begin{aligned}
T_{\text {prl }} & = T_{\text {seq }} = \left\{T\left(x_{\text {test }}\right), T\left(d_1\right), \cdots, T\left(d_N\right)\right\}, \\
A_{\text {prl }} & = [a_{ij}]_{l \times l} \\ &=
    \begin{cases}
      0 & \text{for } 0 \leq j < i < l, \\ 
      0 & \text{between } W_m \text{ and } W_{k}, m \neq k \in [1, \phi] \\
      1 & \text{otherwise}
    \end{cases}, \\
P_{\text {prl }} & =\underbrace{\{0,1, \cdots, p_{\text {max}}\}, \cdots, \{0,1, \cdots, p_{\text {max}}\}}_{\phi \text{ times}}, \\
& \hspace{1.3em} \{p_{\text {max}}+1,\cdots, l-1\}.
\end{aligned}
\end{equation}

%% file: tables/class_table.tex
\begin{table*}
\centering
\renewcommand{\arraystretch}{1.1}
\resizebox{0.95\textwidth}{!}{%
\begin{tabular}{@{}l@{\hspace{3pt}}ccccccc@{}}
\toprule
\multirow{2}{*}{\textbf{Dataset}}     & \multirow{2}{*}{\textbf{\#Labels}} &   \multicolumn{3}{c}{\textbf{LLaMA 7B}}      & \multicolumn{3}{c}{\textbf{Vicuna 13B}}   \\ \cmidrule(l){3-5} \cmidrule(l){6-8}
                &              & \textbf{Seq} & \textbf{PCW} & \textbf{PE}  & \textbf{Seq} & \textbf{PCW} & \textbf{PE} \\ \midrule
RTE          & 2 & \textbf{72.5} \footnotesize{(±3.3)}  & 70.0 \footnotesize{(±4.5)}          & 69.8 \footnotesize{(±4.6)}          & \textbf{77.6} \footnotesize{(±2.2)} & 76.1 \footnotesize{(±1.6)}          & 75.9 \footnotesize{(±1.5)}          \\
CB           & 3 & \textbf{75.6} \footnotesize{(±10.8)} & 70.7 \footnotesize{(±14.1)}         & 70.9 \footnotesize{(±13.8)}         & 79.4 \footnotesize{(±7.9)}          & \textbf{82.7} \footnotesize{(±2.8)} & \textbf{82.7} \footnotesize{(±2.8)} \\
AGNews       & 4 & 86.9 \footnotesize{(±1.8)}           & \textbf{88.0} \footnotesize{(±0.7)} & \textbf{88.0} \footnotesize{(±0.7)} & 81.2 \footnotesize{(±4.3)}          & \textbf{82.5} \footnotesize{(±5.7)} & \textbf{82.5} \footnotesize{(±5.7)} \\
SST5         & 5 & \textbf{47.8} \footnotesize{(±1.0)}  & 47.6 \footnotesize{(±2.0)}          & 47.6 \footnotesize{(±1.9)}          & 49.4 \footnotesize{(±1.0)}          & \textbf{50.0} \footnotesize{(±1.4)} & \textbf{50.0} \footnotesize{(±1.4)} \\
TREC         & 6 & \textbf{82.5} \footnotesize{(±2.6)}  & 73.7 \footnotesize{(±6.1)}          & 73.4 \footnotesize{(±5.9)}          & \textbf{79.3} \footnotesize{(±5.7)} & 64.9 \footnotesize{(±2.3)}          & 66.6 \footnotesize{(±3.1)}          \\ 
DBPedia      & 14 & 90.4 \footnotesize{(±4.7)}           & \textbf{93.8} \footnotesize{(±1.9)} & \textbf{93.8} \footnotesize{(±1.8)} & 93.7 \footnotesize{(±2.9)}          & \textbf{95.5} \footnotesize{(±2.3)} & \textbf{95.5} \footnotesize{(±2.4)} \\\midrule
NLU Scenario & 18 & 79.9 \footnotesize{(±2.8)}           & \textbf{82.5} \footnotesize{(±1.8)} & \textbf{82.5} \footnotesize{(±1.8)} & 79.8 \footnotesize{(±2.0)}          & \textbf{83.7} \footnotesize{(±1.9)} & \textbf{83.7} \footnotesize{(±1.9)} \\
TREC Fine    & 50 & 52.6 \footnotesize{(±10.4)}          & 44.5 \footnotesize{(±8.6)}          & \textbf{55.4} \footnotesize{(±4.9)} & 54.4 \footnotesize{(±4.2)}          & 54.6 \footnotesize{(±5.3)}          & \textbf{55.4} \footnotesize{(±4.9)} \\
NLU Intent  & 68 & \textbf{60.3} \footnotesize{(±2.9)}           & 59.0 \footnotesize{(±3.6)} & 59.0 \footnotesize{(±3.6)} & 65.3 \footnotesize{(±2.4)}          & \textbf{73.6} \footnotesize{(±1.3)} & \textbf{73.6} \footnotesize{(±1.3)} \\
BANKING77   & 77 & 41.8 \footnotesize{(±2.4)}           & \textbf{46.7} \footnotesize{(±2.2)} & 46.6 \footnotesize{(±2.1)}          & 44.8 \footnotesize{(±1.4)}          & \textbf{52.5} \footnotesize{(±2.0)} & 52.4 \footnotesize{(±1.9)}          \\
CLINIC150   & 150 & 62.0 \footnotesize{(±1.6)}           & \textbf{68.0} \footnotesize{(±1.9)} & \textbf{68.0} \footnotesize{(±1.9)} & 64.5 \footnotesize{(±1.6)}          & 68.1 \footnotesize{(±1.8)}          & \textbf{68.2} \footnotesize{(±1.8)} \\   
\bottomrule
\end{tabular}
}
\caption{Results on coarse-grained (\#Labels $\leq$ 15) and fine-grained (\#Labels > 15) classification tasks utilizing three ICL methods: Sequential baseline, Parallel Context Window (PCW)~\cite{ratner2023parallel}, and Parallel Ensemble (PE). We set the number of parallel windows to 3 as it is the best selection according to~\cite{ratner2023parallel}.}
\label{tab:class}

\end{table*}

%% file: 4_experiment.tex
\section{Experiments}
\subsection{Experiment Setup}
\vspace{1mm}

\paragraph{Classification.} 
We perform ICL evaluation on 11 classification datasets spread among diverse domains --- SST5~\cite{socher2013recursive},
CB~\cite{wang2019superglue},
RTE~\cite{bentivogli2009fifth},
BANKING77~\cite{Casanueva2020},
NLU \& NLU Scenario~\cite{Liu2019BenchmarkingNL},
CLINIC150~\cite{larson-etal-2019-evaluation},
AGNews~\cite{zhang2015character},
DBPedia~\cite{zhang2015character},
TREC \& TREC Fine\cite{li-roth-2002-learning}.
The selection of datasets follows PCW~\cite{ratner2023parallel}.

After randomly selecting from the training set as example instances, we calculate results from 10 seed runs. For test samples, we impose a maximum limit of 1000, and in the absence of a validation set, the test set is used. Our evaluation metric is multi-choice accuracy.
For prompt engineering, we follow PCW~\cite{ratner2023parallel} setting. See more details in Appendix~\ref{appendix:prompt}.

\vspace{1mm}

\input{tables/hotpotqa_table}
\paragraph{Reasoning.}
HotpotQA~\cite{yang2018hotpotqa} is a challenging knowledge-intensive multi-hop reasoning task designed for complex reasoning scenarios. Unlike traditional QA tasks, HotpotQA requires LMs to not only locate relevant information from multiple Wikipedia documents but also to understand and connect these pieces of information in a logical and meaningful way. For instance, to answer the question ``What movie starring Nicole Kidman won her an Academy Award'', we will execute Hop 1: Identify the movies in which Nicole Kidman has acted, and then Hop 2: Determine which of these films led to Nicole Kidman winning an Academy Award. By synthesizing these two pieces of information from separate sources, we obtain the final answer ``The Hour''.

We aim for a more advanced setting to evaluate both the knowledge level and reasoning ability leveraging CoT as in ReAct~\cite{yao2023react}, given that current LLaMAs have already achieved performance comparable to PLMs(ranging from 20\% to 30\%) even when they have no access to golden supporting paragraphs\cite{ratner2023parallel}. 

Adhering to the popular CoT evaluation~\cite{yao2023react, wei2022chain}, we manually crafted 18 multi-step thinking trajectories, as creating hundreds of high-quality demonstrations to reach the maximum token length of the language model(2048) is too expensive. 
We select 500 samples from the distractor test set for evaluation. The predictions are generated using greedy decoding at 0 temperature for reproducibility. See more details in Appendix~\ref{appendix:detail}.

\paragraph{Language Models.}
We choose the LLaMA 7B and Vicuna 13B models~\cite{touvron2023llama} for evaluation due to their alignment with human preferences and strong ability to reason. Vicuna 13B is fine-tuned upon LLaMA 13B on user-shared conversations, which achieves nearly 90\% quality of ChatGPT. While LLaMAs employ rotational positional embedding, they still accommodate parallel modifications and can potentially benefit from them, as handling longer texts results in degradation in models with relative positional embeddings~\cite{alibi}.

\subsection{Result Analysis}

\paragraph{PCW is Weighted Sum Ensemble for classification.}

As shown in Table~\ref{tab:class}, the strength of parallel-integrated methods is not universal. They excel mostly in classification tasks featuring many labels, e.g., BANKING77, CLINIC150. 
To identify the underlying cause, we introduce another parallel method, Parallel Ensemble (PE), which directly applies a weighted sum after the test instance's label is predicted using each context window. The weights for each label candidate are determined by the logits of the newly generated tokens, averaged among the sequence. 

We find PCW and PE have similar performances across most tasks, and sometimes PE even slightly outperforms PCW. This suggests that PCW is simply a weighted sum ensemble among all the windows. Coupled with our next finding of impaired reasoning ability caused by parallel windows, we question its viability as a solution for extending the context of LMs.

\input{figures/cot_compare.tex}
\paragraph{PCW deteriorates CoT Reasoning.}
We conducted experiments to explore how parallel windows influence the reasoning chain. HotpotQA, a knowledge-intensive multi-hop reasoning task known for its difficulty, even for models like GPT3.5 and PaLM 540B, merely achieves around 30\% EM accuracy~\cite{yao2023react, shinn2023reflexion}. This makes it an ideal task to detect if language models' performance degrades throughout the reasoning chain. Here we encourage LMs to progressively solve problems utilizing their inherent knowledge through CoT, following ~\cite{yao2023react} to minimize the noises induced by the accuracy and authenticity of provided or retrieved supporting paragraphs.

As illustrated in Table~\ref{tab:hotpotqa}, we notice a significant gap between the Sequential baseline(\# PW = 1) and PCW. 
When exposed to the same number of demonstrations, the raised number of windows implies sparser attention, resulting in worse performance because the repetitive positional embeddings might confuse the LM. Even when comparing 6-shots with 12- or 18-shots that offer double or triple the examples, the parallel method still falls short. 

Further error analysis depicted in Figure~\ref{fig2} reveals that PCW easily misinterprets the basic logical relation between contexts, sometimes even disregards the question, and provides unrelated answers. None-reasoning error is mainly caused by hallucination, which is less relevant to the rationality of CoT reasoning. Other includes the generation of repetitive sentences or meaningless symbols.

%% file: tables/hotpotqa_table.tex
\begin{table*}[htbp]
\centering
\renewcommand{\arraystretch}{1.1}
\renewcommand\tabcolsep{4.0pt}
\begin{tabular}{@{}cp{2cm}<{\centering}p{2cm}<{\centering}p{2cm}<{\centering}p{2cm}<{\centering}p{2cm}<{\centering}p{2cm}<{\centering}@{}}
\toprule
\multirow{2}{*}{\textbf{\#Shots}} & \multicolumn{3}{c}{\textbf{LLaMA   7B}}                      & \multicolumn{3}{c}{\textbf{Vicuna   13B}}                    \\ \cmidrule(l){2-4}  \cmidrule(l){5-7}
                                   & \textbf{\#PW = 1} \footnotesize{(Sequential)} & \textbf{\#PW = 2} & \textbf{\#PW = 3} & \textbf{\#PW = 1} \footnotesize{(Sequential)} & \textbf{\#PW = 2} & \textbf{\#PW = 3} \\ \midrule
2                                  &\textbf{14.53} \footnotesize{(±2.74)}      & 0.07 \footnotesize{(±0.09)}              & -                  & \textbf{16.93} \footnotesize{(±3.79)}      & 0.20 \footnotesize{(±0.16)}                  & -                  \\
3                                  & \textbf{18.60} \footnotesize{(±1.50)}        & -                  & 0.67 \footnotesize{(±0.81)}                  & \textbf{23.00} \footnotesize{(±1.45)}       & -                  & 3.33 \footnotesize{(±3.17)}                  \\
6                                  & \textbf{20.13} \footnotesize{(±1.11)}      & 19.13 \footnotesize{(±1.00)}       & 16.27 \footnotesize{(±0.57)}      & \textbf{23.60} \footnotesize{(±0.49)}       & 23.40 \footnotesize{(±1.61)}       & 22.47 \footnotesize{(±0.81)}      \\
12                                 & \textbf{19.87} \footnotesize{(±0.25)}      & 19.07 \footnotesize{(±0.74)}      & 18.27 \footnotesize{(±0.74)}      & \textbf{24.07} \footnotesize{(±0.84)}      & 23.07 \footnotesize{(±0.34)}      & 22.80 \footnotesize{(±0.00)}        \\
18                                 & \textbf{20.27} \footnotesize{(±0.82)}      & 19.53 \footnotesize{(±1.27)}      & 18.27 \footnotesize{(±0.34)}      & \textbf{24.60} \footnotesize{(±1.30)}        & 24.10 \footnotesize{(±0.75)}       & 22.80 \footnotesize{(±1.14)}       \\ \bottomrule
\end{tabular}%
\caption{CoT results on HotpotQA evaluated in Exact Match score. \#PW denotes the number of parallel windows, higher PW means finer-grained windows, and \#PW = 1 demonstrates the sequential baseline.}
\label{tab:hotpotqa}
\end{table*}

%% file: figures/cot_compare.tex
\begin{figure}[]
    \centering
    \includegraphics[width=\linewidth]{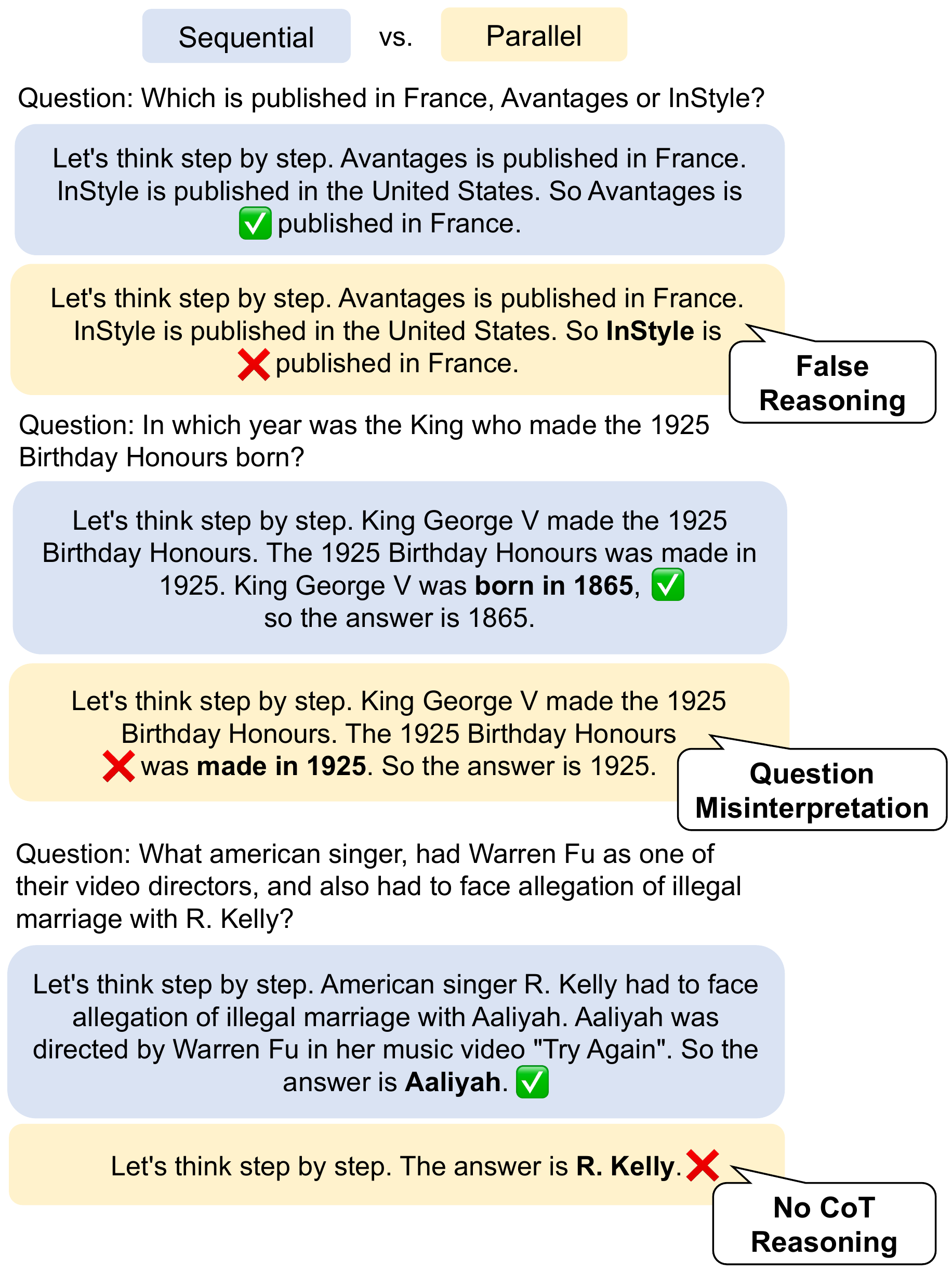}
    \vspace{-1mm}
    \caption{Case study on closed-book HotpotQA CoT reasoning, where the sequential method succeeds but PCW fails in the reasoning due to reasons above.}
    \label{fig2}
    \vspace{-5mm}
\end{figure}

%% file: 6_conclusion.tex
\section{Conclusion}
We raise concerns about the use of parallel-integrated methods to address context length restriction: (1) PCW is functionally equal with a simple weighted sum ensemble on label distribution among context windows; (2) PCW degrades the multi-step reasoning capabilities of LLMs in complex tasks requiring knowledge understanding. Despite the fact that parallel-integrated methods sometimes show better classification performance when the label space is large, they merely brute-force ensemble each window's context, consequently weakening logical reasoning and knowledge comprehension.

%% file: appendix.tex
\section{Appendix}

\subsection{Prompts}
\input{tables/appendix_pcw}
\subsubsection{Reasoning}
We manually write 18 Chain-of-Thoughts demonstrations for the HotpotQA task including two subcategories --- comparison and bridge. In bridge reasoning, the answer to the question requires making a connection between two or more pieces of information that are not directly related. The model needs to ``bridge'' the gap between these pieces of information in order to arrive at the correct answer. Comparison reasoning involves comparing two or more entities based on their attributes or related facts. This requires the model to understand and compare information from different facts. 
They are selected from the distractor test set while ensuring no overlap with the evaluation data pool.
See Table~\ref{tab:hotpot_prompt} for details. 

\subsubsection{Classification}\label{appendix:prompt}
We strictly follow the prompting from ~\cite{ratner2023parallel} in order to make a fair comparison. Therefore, we encourage a read of the original paper for details.

\input{tables/prompt_hotpot}

\subsection{Supplementary Results}
We evaluate the most fine-grained parallel window method, i.e., PCW Single, where the window span is 1. We find that under such conditions, the parallel method drastically declines due to excessive repetition of positional embeddings in context windows, as shown in Table~\ref{tab:single_pcw}.
We choose $n_{max}$ for each dataset to be the shot number that fills in the maximum token length of LMs, i.e., 2048 for Vicuna. We set the window size as 3 to align with the main results in Section 3.

It is evident that as the number of parallel windows increases, there is a dramatic drop in In-Context Learning (ICL) performance. This decline is especially notable in datasets such as BANKING77 and CLINIC150, which contain more than 50 labels. This is because of a prediction bias favoring one certain label. Above results demonstrate the negative effects of repeated positional embeddings for language models.

\subsection{Experiment Details}\label{appendix:detail}
We use LLaMa 7B and Vicuna 13B v1.1 checkpoint from HuggingFace for evaluation. To accelerate inference time, we adopt the int8 quantization for the language models.

For classification tasks, we sample 10 times from the training set, limiting the maximum test samples to 1000. We record the mean and variance for each seed run across all experimental results. Figure~\ref{fig1} above shows LLaMA 7B results.
For the reasoning task, we sample from the manually designed demonstration pool with 3 seeds, restricting the size of the test samples to 500. We randomly select 100 samples to derive Table~\ref{tab:error}. Figure~\ref{fig1} below shows Vicuna 13B results.

%% file: tables/appendix_pcw.tex
\begin{table*}[b]
\centering
\renewcommand{\arraystretch}{1.05}
\begin{tabular}{@{}lcccc@{}}
\toprule
\textbf{Method}       & Seq         & PCW         & PCW Single   & PCW Single   \\ \midrule
\textbf{\# Shots}     & $n_{max}$           & $3 * n_{max}$     & $n_{max}$            & $3 * n_{max}$        \\
\textbf{RTE}          & 77.6 (±2.2) & 76.1 (±1.6) & 59.3 (±5.1)  & 56.2 (±1.7)  \\
\textbf{CB}           & 79.4 (±7.9) & 82.7 (±2.8) & 66.6 (±13.4) & 55.1 (±9.8)  \\
\textbf{AGNews}       & 81.2 (±4.3) & 82.5 (±5.7) & 86.5 (±1.6)  & 69.2 (±15.8) \\
\textbf{SST5}         & 49.4 (±1.0) & 50.0 (±1.4) & 29.4 (±2.8)  & 26.2 (±1.5)  \\
\textbf{TREC}         & 79.3 (±5.7) & 64.9 (±2.3) & 20.1 (±3.5)  & 18.7 (±1.2)  \\
\textbf{DBPedia}      & 93.7 (±2.9) & 95.5 (±2.3) & 89.9 (±2.7)  & 82.1 (±5.1)  \\
\textbf{NLU Scenario} & 79.8 (±2.0) & 83.7 (±1.9) & 9.4 (±1.9)   & 2.4 (±1.3)   \\
\textbf{TREC Fine}    & 54.4 (±4.2) & 54.6 (±5.3) & 8.2 (±1.9)   & 9.9 (±1.8)   \\
\textbf{NLU Intent}   & 65.3 (±2.4) & 73.6 (±1.3) & 3.6 (±1.0)   & 3.4 (±0.5)   \\
\textbf{BANKING77}    & 44.8 (±1.4) & 52.5 (±2.0) & 8.5 (±5.4)   & 1.6 (±0.3)   \\
\textbf{CLINIC150}    & 64.5 (±1.6) & 68.1 (±1.8) & 2.0 (±0.4)   & 0.4 (±0.2)   \\ \bottomrule
\end{tabular}
\caption{Supplementary results on ICL classification tasks for Vicuna 13B. }
\label{tab:single_pcw}
\end{table*}

%% file: tables/prompt_hotpot.tex
\begin{table*}[htbp]
\caption{CoT demonstrations for HotpotQA}
    \centering
    \scalebox{0.95}{
    \begin{tabular}{@{} p{2cm} p{13cm} @{}}
        \toprule
        \texttt{Question} & \texttt{What is the elevation range for the area that the eastern sector of the Colorado orogeny extends into?} \\
        \texttt{Thought} & \texttt{The eastern sector of Colorado orogeny extends into the High Plains. High Plains rise in elevation from around 1,800 to 7,000 ft, so the answer is 1,800 to 7,000 ft.} \\
        \texttt{Answer} & \texttt{1,800 to 7,000 ft} \\
        \texttt{Question} & \texttt{Musician and satirist Allie Goertz wrote a song about the "The Simpsons" character Milhouse, who Matt Groening named after who?} \\
        \texttt{Thought} & \texttt{The character Milhouse was named after U.S. president Richard Nixon, so the answer is Richard Nixon.} \\
        \texttt{Answer} & \texttt{Richard Nixon} \\
        \texttt{Question} & \texttt{Which documentary is about Finnish rock groups, Adam Clayton Powell or The Saimaa Gesture?} \\
        \texttt{Thought} & \texttt{Clayton Powell (film) is a documentary about an African-American politician, not Finnish rock groups. So the documentary about Finnish rock groups must instead be The Saimaa Gesture.} \\
        \texttt{Answer} & \texttt{The Saimaa Gesture} \\
        \texttt{Question} & \texttt{What profession does Nicholas Ray and Elia Kazan have in common?} \\
        \texttt{Thought} & \texttt{Professions of Nicholas Ray are director, screenwriter, and actor. Professions of Elia Kazan are director, producer, screenwriter, and actor. So profession Nicholas Ray and Elia Kazan have in common is director, screenwriter, and actor.} \\
        \texttt{Answer} & \texttt{director, screenwriter, actor} \\
        \texttt{Question} & \texttt{Which magazine was started first Arthur's Magazine or First for Women?} \\
        \texttt{Thought} & \texttt{Arthur's Magazine was started in 1844. First for Women was started in 1989. 1844 (Arthur's Magazine) < 1989 (First for Women), so Arthur's Magazine was started first.} \\
        \texttt{Answer} & \texttt{Arthur's Magazine} \\
        \texttt{Question} & \texttt{Were Pavel Urysohn and Leonid Levin known for the same type of work?} \\
        \texttt{Thought} & \texttt{Pavel Urysohn is a mathematician. Leonid Levin is a mathematician and computer scientist. So Pavel Urysohn and Leonid Levin have the same type of work.} \\
        \texttt{Answer} & \texttt{Yes} \\
        \texttt{Question} & \texttt{Which of Jonny Craig and Pete Doherty has been a member of more bands?} \\
        \texttt{Thought} & \texttt{Jonny Craig has been a member of four bands, Dance Gavin Dance, Emarosa, Rain City Drive, Isles, and Glacierswhile. Pete Doherty has been a member of three bands, Babyshambles, Peter Doherty and the Puta Madres. So Jonny Craig has been a member of more bands.} \\
        \texttt{Answer} & \texttt{Jonny Craig} \\
        \texttt{Question} & \texttt{If You Ever Get Lonely was covered by what Lyric Street Records-affiliated band?} \\
        \texttt{Thought} & \texttt{If You Ever Get Lonely was covered by American country music duo Love and Theft, which is a Lyric Street Records-affiliated band, so the answer is Love and Theft.} \\
        \texttt{Answer} & \texttt{Love and Theft} \\
        \texttt{Question} & \texttt{Jaclyn Stapp is married to the former frontman of a band that disbanded in what year?} \\
        \texttt{Thought} & \texttt{Jaclyn Stapp is married to Scott Stapp, the voice of the band Creed. Creed was an American rock band from Tallahassee, Florida, active from 1994 to 2004. So Creed disbanded in  2004.} \\
        \texttt{Answer} & \texttt{2004} \\
    \end{tabular}
}
\label{tab:hotpot_prompt}
\end{table*}

\begin{table*}[htbp]
    \centering
    \scalebox{0.95}{
    \begin{tabular}{@{} p{2cm} p{13cm} @{}}
        \texttt{Question} & \texttt{The W. H. Shipman House is in what Hawaii county?} \\
        \texttt{Thought} & \texttt{W. H. Shipman House is located at 141 Kaiulani Street, Hilo. Hawaii County, Hawaii is the county that Hilo is in, so the answer is Hawaii County.} \\
        \texttt{Answer} & \texttt{Hawaii County} \\
        \texttt{Question} & \texttt{The trophy given to the winner of the college football game between University of Idaho Vandals and University of Montana Grizzlies is derived from which European culture?} \\
        \texttt{Thought} & \texttt{The trophy given to the winner of the college football game between University of Idaho Vandals and University of Montana Grizzlies is the Little Brown Stein. The Little Brown Stein is a large stein mug, which is derived from German culture.} \\
        \texttt{Answer} & \texttt{German} \\
        \texttt{Question} & \texttt{In what year was the singer who popularized "I'll Keep the Lovelight Burning" born?} \\
        \texttt{Thought} & \texttt{"I'll Keep the Lovelight Burning (In My Heart)" is a popular song  written by Bennie Benjamin and George David Weiss,  popularized by Patti Page in 1949. Patti Page was born in 1927, so the answer is 1927.} \\
        \texttt{Answer} & \texttt{1927} \\
        \texttt{Question} & \texttt{Stronger Together was used for the campaign comprised of the Vice Presidential candidate who was a Senator from what state?} \\
        \texttt{Thought} & \texttt{Tim Kaine was a Senator from Virginia who used Stronger Together for the campaign, so the answer is Virginia.} \\
        \texttt{Answer} & \texttt{Virginia} \\
        \texttt{Question} & \texttt{Which facility was founded in Missouri, Discovery Zone or Valentino's?} \\
        \texttt{Thought} & \texttt{Valentino's was founded in Lincoln, Nebraska in 1957. 1989 (Discovery Zone) < 1957 (Valentino's), so Discovery Zone was founded in Missouri first.} \\
        \texttt{Answer} & \texttt{Discovery Zone} \\
        \texttt{Question} & \texttt{Kam Heskin plays Paige Morgan in a 2004 film directed by who?} \\
        \texttt{Thought} & \texttt{Kam Heskin played Paige Morgan in The Prince and Me film franchise. The Prince and Me film franchise was directed by Martha Coolidge in 2004. So the answer is Martha Coolidge.} \\
        \texttt{Answer} & \texttt{Martha Coolidge} \\
        \texttt{Question} & \texttt{Otterington railway station was on a 393 mi major railway that linked Edinburgh to what city?} \\
        \texttt{Thought} & \texttt{Otterington railway station was on the East Coast Main Line. The East Coast Main Line linked Edinburgh to London King's Cross station. So the answer is London.} \\
        \texttt{Answer} & \texttt{London} \\
        \texttt{Question} & \texttt{Who wrote the book that inspired the name of the CD/DVD "Here at the End of All Things"?} \\
        \texttt{Thought} & \texttt{The title comes from a quote from the book The Lord of the Rings: The Return of the King. The Lord of the Rings was written by J. R. R. Tolkien, so the answer is J. R. R. Tolkien.} \\
        \texttt{Answer} & \texttt{J. R. R. Tolkien} \\
        \texttt{Question} & \texttt{Are both Lygodium or Maxillaria a genus of orchids?} \\
        \texttt{Thought} & \texttt{Maxillaria is a genus of orchids. Lygodium is not a genus of orchids. So the answer is No.} \\
        \texttt{Answer} & \texttt{No} \\
        \bottomrule
    \end{tabular}
    }
\end{table*}